\documentclass[runningheads]{llncs}
\usepackage{graphicx}
\usepackage[english]{babel}
\usepackage{caption}
\usepackage[numbers]{natbib}
\usepackage{tabto}
\usepackage{array}
\usepackage{soul}
\usepackage{mathtools,xparse}
\DeclarePairedDelimiter{\norm}{\lVert}{\rVert}
\graphicspath{ {pics/} }
\begin{document}
\title{The Coherent Point Drift for Clustered Point Sets}
%
%

\author{Dmitry Lachinov \inst{1}\orcidID{0000-0002-2880-2887} \and
Vadim Turlapov\inst{1}\orcidID{0000-0001-8484-0565}}

\authorrunning{D. Lachinov and V. Turlapov}

%

\institute{Lobachevsky State University, Gagarina ave. 23, 603950 Nizhny Novgorod, Russian Federation 
\email{\\ \{dlachinov, vadim.turlapov\}@gmail.com }}

\maketitle              
\begin{abstract}
The problem of non-rigid point set registration is a key problem for many computer vision tasks. In many cases the nature of the data or capabilities of the point detection algorithms can give us some prior information on point sets distribution. In non-rigid case this information is able to drastically improve registration results by limiting number of possible solutions. In this paper we explore use of prior information about point sets clustering, such information can be obtained with preliminary segmentation. We extend existing probabilistic framework for fitting two level Gaussian mixture model and derive closed form solution for maximization step of the EM algorithm. This enables us to improve method accuracy with almost no performance loss.
We evaluate our approach and compare the Cluster Coherent Point Drift with other existing non-rigid point set registration methods and show it’s advantages for digital medicine tasks, especially for heart template model personalization using patient’s medical data.

\end{abstract}
\section{Introduction}
Points set registration task often arises in many computer vision problems. Statistical shape reconstruction, medical image registration, template personalization and various number of other tasks incorporate point set registration.
This task consists of computing point correspondences and spatial transformation between two given point sets.
These sets could be artificially generated, obtained with scanner or extracted from an image using various point detection approaches. In practice point set registration is primarily used for obtaining transformation between pre-defined template and points extracted from some data. In this paper we consider a special case of point set registration - where information on point sets clustering is already known. As an example, points extracted from CT scan of the person's heart can have such property. Heart ventricles and atriums are considered as clusters in this case.
\newline
The task of non-rigid point set registration is ill-posed by itself. So additional constrains are used to achieve a solution. In general, the type of the transformation between template set and extracted set is unknown, so the solution of the problem results in finding optimal function with respect to problem's constrains in non-rigid transformations class. These constrains are motivated by assumptions on point sets distribution and target transformation type.
Desired solution should describe original transformation the best, be robust to outliers, noise and missing data, be able to process high dimensional data and have reasonable computational complexity.
The Coherent Point Drift \cite{cpd2010} stands out of other methods\cite{review} and is able to provide accurate results in reasonable time. Still, it fails to perform in some complex cases. In this paper we extend it's capabilities by taking the idea of including prior information described by V. Golyanik et al. \cite{ecpd2016} and expand it further to include information on point sets clustering. This information often comes naturally with properties of the point extraction algorithm or nature of the data. For example, many biomedical tasks include segmentation as pre-processing tool but no information on clusters is actually used.
In this paper we present CPD modification for clustered point sets that extends existing approach to adopt prior information.
In the experiments section we evaluate proposed method using 20 CT samples from MMWHS-2017 \cite{mmwhs1,mmwhs2,mmwhs3} challenge and compare results with CPD\cite{cpd2010} and ECPD \cite{ecpd2016}. For the results evaluation we are using Hausdorff distance \cite{hausdorff} and it's variant for clustered point sets. This shows advantages of the proposed method over existing ones for the given task.
\newline

\section{Related Work}
The Coherent Point drift algorithm as a working idea was formulated at 2007 by Myronenko et al. \cite{cpd2007} for the first time. The idea behind the method is to formulate point set registration as a maximum likelihood estimation problem. The transformation (model parameters) aligning point set Y with point set X that maximizes likelihood with subject to constrains is chosen as the result. Only non-rigid registration case is considered in that work. The authors expand this approach further at \cite{cpd2010} for both rigid and non-rigid cases. Techniques reducing algorithm's complexity and significantly improving performance were proposed. Authors demonstrate method's advantages over existing approaches like LM-ICP \cite{lmicp} and TPS-RPM \cite{tpsrpm} in presence of noise and outliers. 
Review of the method's modern applications is given in \cite{reviewour}. \newline
Golyanik et al. extended existing approach to take prior knowledge into account \cite{ecpd2016}. In their work they adopted coarse-to-fine strategy for processing large point set in reasonable time. Authors considered a special case of the non-rigid registration problem, when a sparse set of correspondence priors is known in advance. A modification for Coherent Point Drift was proposed that allows to embed correspondence priors in a closed-form.
ECPD employs correspondence preserving subsampling counterbalancing the polynomial complexity by splitting the problem into two subproblems of smaller size and reducing the number of operations by a linear factor.
\newline
In \cite{gcpd}, two novel elements were proposed: (1) generalization of the CPD (gCPD) that used a procedure of group-wise registration (alignment) for statistical shape reconstruction and (2) more complex point sets, in which each point that belonged to the surface was associated with a normal; moreover, instead of the Gaussian mixture model (GMM), the hybrid mixture model (HMM) was employed that combined the Student t-distribution and the von Mises–Fisher distribution for a point and a normal, respectively. Experiments used clinical data on 27 brain ventricles (Neuro) and 15 hearts (Cardiac). The HMM was shown to provide a significant increase in
accuracy compared to the methods (including the gCPD) that used conventional point sets. As a metric,
the mean surface distance (MSD) was employed.
\subsection{Coherent Point Drift}
The Coherent point drift algorithm was introduced in \cite{cpd2007} and improved later in \cite{cpd2010}. This method is designed to align two point sets $X$ and $Y$ where one point set represents GMM centroids and another represents data. $X$ points set representing the data is considered to be fixed, $Y$ is moving. In order to make $Y$ points move coherently regularization is used.\newline
Authors are using the following notation we will also stick to.
\begin{itemize}
\item $D$ - dimension of the point sets;
\item $N, M$ - number of points in X and Y respectively;
\item $X_{NxD = (x_{1}, ..., x_{N})^{T}}$ - the first point set representing the data;
\item $Y_{MxD = (y_{1}, ..., y_{M})^{T}}$ - the second point set representing the centroids;
\item $T(Y,\Theta)$ - transformation $T$ applied to $Y$, where $\Theta$ is the transformation parameters;
\item $I$ - identity matrix;
\item $1$ - column vector of all ones;
\item $d(a)$ - diagonal matrix formed from the vector a.
\end{itemize}
The $Y$ set is GMM centroids with probability density written as
$$p(x) = \sum_{m=1}^{M+1}{P(m)p(x|m)}$$
where $p(x|m) = \frac{1}{{(2\pi\sigma^{2})}^{\frac{D}{2}}}exp^{-\frac{\norm{x-y_{m}}^{2}}{2\sigma^{2}}}$. To account presence of the noise and outliers $P(x|M+1) = 1/N$ was added to mixture model. Equal isotropic covariances $\sigma^2$ and equal membership probabilities $P(m) = 1/M$ are used. Resulting mixture model density can be written as $$p(x) = \frac{\omega}{N} + (1-\omega)\sum_{m=1}^{M+1}{p(x|m)}$$ where $\omega$ is the fraction of the noise in the model.\newline
The model parameters $\Theta$ are estimated by maximizing likelihood function. This task is equivalent to minimization of the negative log-likelihood $E(\Theta,\sigma^{2})$ that can be written as
$$E(\Theta,\sigma^{2}) = -\sum_{n=1}^{N}{log(p(x_n))}$$
For parameters estimation EM algorithm \cite{bishop} is used. It consists of two steps, first, posterior probabilities $p(m|x_n)$ are calculated using old parameters, then new parameters values are estimated during second step. The upper bound of the negative log-likelihood can be written as
$$Q = - \sum_{n=1}^{N}{ \sum_{m=1}^{M+1}{P^{old}(m|x_n)log(P^{new}(m)p^{new}(x_n|m))}}$$
rewriting it we can get the following:
$$Q(\Theta,\sigma^2) = \frac{1}{2\sigma^2}\sum_{n=1}^{N}{ \sum_{m=1}^{M}{P^{old}(m|x_n)\norm{x_n - T(y_m,\Theta)}^2}} + \frac{N_pD}{2}log\sigma^2$$
where $N_p = \sum_{n=1}^{N}{ \sum_{m=1}^{M+1}{P^{old}(m|x_n)}} \leq N$ and $P^{old}$ - posterior probabilities of mixture components calculated with old parameters values. $P^{old}$ can be written as
$$p_{mn} = \frac{exp^{-\frac{\norm{x_n-T(y_m,\Theta)}^2}{2\sigma^2}}}{\sum_{k=1}^{M}{exp^{-\frac{\norm{x_n-T(y_k,\Theta)}^2}{2\sigma^2}} + c}}$$
where $c = \frac{{(2\pi\sigma^2)}^\frac{D}{2}\omega}{1-\omega} \frac{M}{N}$. Minimizing the function $Q$, we necessarily decrease the negative log-likelihood function $E$, unless it is already at a local minimum.
In the paper \cite{cpd2010} both rigid and non-rigid cases are considered. From biomedical applications perspective non-rigid case is more valuable, so we will focus on non-rigid CPD version.
In order to represent non-rigid transformation the displacement field $v$ is introduced.
$$T(Y) = Y + v(Y)$$
Adding regularization term to negative log-likelihood we obtain
$$f(v,\sigma^2) = E(v,\sigma^2) + \frac{\lambda}{2}\phi(v)$$
where $E$ is negative log-likelihood and $\phi$ is regularization term.
Authors show that regularization term can be rewritten as
$$\phi(v)=\norm{v}_{H_m}^2 = \norm{Pv}^2$$ where operator $P$ extracts the high frequency content \cite{cpd2010, reg1, reg2}.
Authors show that displacement function $v(z)$ can be written as
$$v(z) = \sum_{m=1}^{M}{w_mG(z,y_m)}$$
where $w_m=\frac{1}{\lambda\sigma^2}\sum_{n=1}^{N}{P^{old}(m|x_n)(x_n - (y_m+v(y_m))}$ and $G(z,y) = e^{-\frac{1}{2\beta^2}\norm{z-y}^2}$.
Matrix $W = (w_1,...,w_m)^T$  can be evaluated by solving following equation
$$(G + \lambda\sigma^2d(P1)^{-1})W=d(P1)^{-1}PX-Y$$
where $g_{i,j} = G(y_i,y_j)$ and 
\begin{equation}
\label{eq:pmn}
p_{mn} = \frac{exp^{-\frac{\norm{x_n-T(y_m,\Theta)}^2}{2\sigma^2}}}{\sum_{k=1}^{M}{exp^{-\frac{\norm{x_n-T(y_k,\Theta)}^2}{2\sigma^2}} + \frac{{(2\pi\sigma^2)}^\frac{D}{2}\omega}{1-\omega} \frac{M}{N} }}
\end{equation}
\subsection{Extended Coherent Point Drift}
Extended Coherent Point Drift \cite{ecpd2016} integrates prior knowledge into registration algorithm. It requires sparse set of points correspondences between point sets to be defined.\newline
These correspondences are defined as pairs $(x_j,y_k)$ where $(j,k) \in N_c \subseteq N^2$. Correspondence priors are modeled as a product of particular independent density functions
$$P_c(N_c) = \prod_{(j,k)\in N_C}{}p_c(x_j,y_k)$$
where $p_c(x_j,y_k) = \frac{1}{{2\pi\alpha^{2}}^{\frac{D}{2}}}exp^{-\frac{\norm{x_j-T(y_k,\Theta)}^{2}}{2\alpha^{2}}}$ and $\alpha > 0$ is priors' degree of reliability. Correspondence priors are incorporated into GMM
$$\Tilde{p}(x)=P_c(N_c)p(x)$$
Modified negative log-likelihood will be written as
$$\Tilde{E}(\Theta, \sigma^2)= E(\Theta, \sigma^2) + \sum_{(j,k)\in N_c}{log(p_c(x_j,y_k))}$$
Upper bound $Q$ will be defined as
$$\Tilde{Q} = Q + \frac{1}{2\alpha^2}\sum_{n=1}^{N}{ \sum_{m=1}^{M}{\Tilde{P}_{mn}\norm{x_n - T(y_m,\Theta)}^2}}$$
where $\Tilde{P}_{mn}$ defined as
$$\Tilde{p}_{mn} = \begin{cases} 1, & \mbox{if } (m,n)\in N_c \\ 0, & \mbox{otherwise } \end{cases}$$
$\Tilde{P}_{mn}$ matrix can be pre-calculated.
For non-rigid case upper-bound function $\Tilde{Q}$ will be written as
\begin{align*}
\Tilde{Q}(\Theta,\sigma^2) = &\frac{1}{2\sigma^2}\sum_{n=1}^{N}{ \sum_{m=1}^{M}{P^{old}(m|x_n)\norm{x_n - T(y_m,\Theta)}^2}} +
\frac{1}{2\alpha^2}\sum_{n=1}^{N}{ \sum_{m=1}^{M}{\Tilde{P}_{mn}\norm{x_n - T(y_m,\Theta)}^2}} +\\
&\frac{N_pD}{2}log\sigma^2 +
\frac{\lambda}{2}\phi(v)
\end{align*}
The following linear system is solved during maximization step of the EM algorithm
$$(d(P1)G + \frac{\sigma^2}{\alpha^2}d(\Tilde{P}1)G + \lambda\sigma^2I)W = 
PX - d(P1)Y + \frac{\sigma^2}{\alpha^2}(\Tilde{P}X - d(\Tilde{P}1)Y)$$
\section{Proposed method}
In this paper we consider a special case for point set registration, where point sets are clustered and correspondences between clusters of the different sets are known. These points can be extracted form heart CT scans so clusters will represent different heart cavities. In general, such information can be extracted for various number of biomedical tasks using different segmentation techniques like neural nets \cite{dlreview}. 
In the paper we will be using following notation:
\begin{itemize}
\item $c \in N$ - number of the cluster, $C$ clusters in total;
\item $P(c|x_n,m)$ - probability of $x_n$ and $y_m$ both to be inside cluster $c$;
\item $P(c|x_n)$ - probability of $x_n$ to be inside cluster $c$;
\item $P(c|m)$ - probability of $y_m$ to be inside cluster $c$;
\item $P(c)$ - probability of the cluster $c$;
\end{itemize}
Let us define additional cluster $C+1$ so that $P(C+1|x_n,m) = 1 -\sum_{c=1}^{C}{P(c|x_n,m)}$ with $P(C+1)=0$. This cluster includes all pairs $(x_n,y_m)$ that don't share any cluster. 
We will rewrite probability density function of the mixture model as
$$\Bar{p}(x)=\sum_{c=1}^{C+1}{P(c) \sum_{m=1}^{M+1}{p(x|m,c)P(m|c)}}$$
The negative log-likelihood is defined as
$$\Bar{E} = - \sum_{n=1}^{N}{\Bar{p}(x)}$$
And it's upper bound can be written as
$$\Bar{Q} = - \sum_{c=1}^{C+1}{ P(c) \sum_{n=1}^{N}{ \sum_{m=1}^{M+1}{ P^{old}(m|x_n,c) log(p^{new}(m|x_n)p^{new}(m|c) ) }}}$$

$$\Bar{Q}(\Theta,\sigma^2) = \frac{1}{\sigma^2} \sum_{c=1}^{C+1}{ P(c) \sum_{n=1}^{N}{ \sum_{m=1}^{M}{ P^{old}(m|x_n,c)
\norm{x_n - T(y_m,\Theta)}^2 }}} + \frac{\Bar{N}_pD}{2}log(\sigma^2)$$
where $\Bar{N}_p = \sum_{c=1}^{C+1}{ P(c) \sum_{n=1}^{N}{ \sum_{m=1}^{M}{ P^{old}(m|x_n,c)}}}$.
The posterior probabilities $P(m|x_n,c)$ can be calculated as follows
$$P(m|x_n,c) = \frac{P(c|x_n,m)p(x|m)}{ \sum_{k=1}^{M}{P(c|x_n,m=k)p(x|m=k)} }$$
After adding regularization and removing the term that corresponds to $P(C+1)$ we can write upper bound as

$$\Bar{Q}(W) = \frac{1}{\sigma^2} \sum_{c=1}^{C}{ P(c) \sum_{n=1}^{N}{ \sum_{m=1}^{M}{ P^{old}(m|x_n,c)
\norm{x_n - (y_m + G(m,)W)}^2 }}} + \frac{\lambda}{2}tr(W^TGW)$$
After rewriting equations in matrix form and setting derivative to zero we will get
$$\frac{\partial \Bar{Q}(W)}{\partial W} = \frac{C}{\sigma^2}\sum_{c=1}^{C}{ P(c) (d(\Bar{P}(c)1)(Y+GW)  - \Bar{P}(c)X)  }+ \lambda G W = 0$$
So during maximization step we iteratively update matrix $W$ using following equation:
$$(\sum_{c=1}^{C}{P(c) d(\Bar{P}(c)1 )W} +\lambda\sigma^2I )W = \sum_{c=1}^{C}{P(c)( \Bar{P}(c)X - d(\Bar{P}(c)1)Y)}$$
where $\Bar{P}(c) = P^{old}(m|x_n,c)$.
Setting derivatives to zero we can get $\sigma^2$ update rule
$$\sigma^2 = \frac{1}{\Bar{N}_pD}(
tr(X^T(\sum_{c=1}^{C}{P(c)d(\Bar{P}(c)1)})X) - 
2tr(((\sum_{c=1}^{C}{P(c)\Bar{P}(c)})X)^TT) +
tr(T^Td(\sum_{c=1}^{C}{P(c)\Bar{P}(c)1})T)
)$$
where $T = Y+GW$\newline
Strictly speaking $P(c|x_n,m)$ is unknown, but it has to be initialized. From it's definition - probability of given $x_n$ and $y_m$ to be in the same cluster $c$;
we can estimate it with product $P(c|x_n,m) = \frac{1}{NM}\sum_{n=1}^{N}{I(x_n \in c)} \sum_{m=1}^{M}{I(y_m \in c)}$, where $I(x_n \in c)$ equals to 1 if $x_n$ belongs to cluster $c$ and equals to 0 otherwise.
\newline\newline
As a result, Cluster Coherent Point drift can be described as
\begin{itemize}
\item \textbf{inputs}: point sets $X$ and $Y$, probabilities $P(c|x_n)$ and $P(c|y_m)$;
\item \textbf{outputs}: aligned point set T(Y);
\item \textbf{algorithm}: CCPD;
\item Initialize: \begin{align*}
&\beta > 0, \lambda > 0, 0 \leq \omega \leq 1, 
\sigma^2 = \frac{1}{DMN}\sum_{n=1}^{N}{ \sum_{m=1}^{M}{\norm{x_n-y_m}^2}}. \\
&G(y_i,y_j) = e^{-\frac{1}{2\beta^2}\norm{y_i-y)j}^2}, P(c|x_n,m) = \frac{1}{NM}\sum_{n=1}^{N}{I(x_n \in c)} \sum_{m=1}^{M}{I(y_m \in c)};
\end{align*}
\item Repeat until convergence;
\begin{itemize}
    \item E-step: compute
    $$P(m|x_n,c) = \frac{P(c|x_n,m)p(x|m)}{ \sum_{k=1}^{M}{P(c|x_n,m=k)p(x|m=k)} }$$
    \item M-step: solve
    $$(\sum_{c=1}^{C}{P(c) d(\Bar{P}(c)1 )W} +\lambda\sigma^2I )W = \sum_{c=1}^{C}{P(c)( \Bar{P}(c)X - d(\Bar{P}(c)1)Y)}$$
    \item update $T, \Bar{N}_p, \sigma^2$
\end{itemize}
\item Aligned set is $T=Y+GW$
\end{itemize}
The key difference between ECPD and proposed method is the assumption made about prior information.
\section{Evaluation}
\begin{figure}[h]
\includegraphics[scale=0.26]{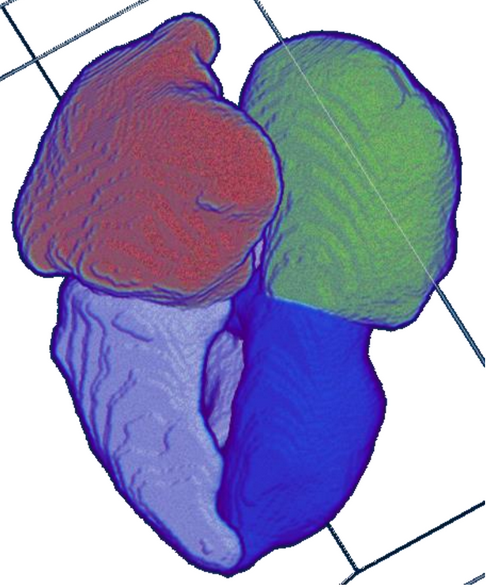}
\centering
\caption{Example of the heart cavities labels from MMHWS dataset. Different colors represent different classes. Dataset contains more labels than represented on the figure.}
\label{fig:mmwhs}
\end{figure}

\begin{figure}[h]
\includegraphics[scale=0.42]{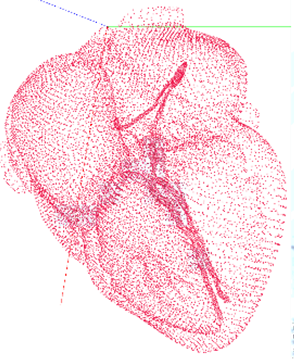}
\centering
\caption{Example of the heart model personalization}
\label{fig:personalization}
\end{figure}

\begin{figure}[h]
\includegraphics[scale=0.42]{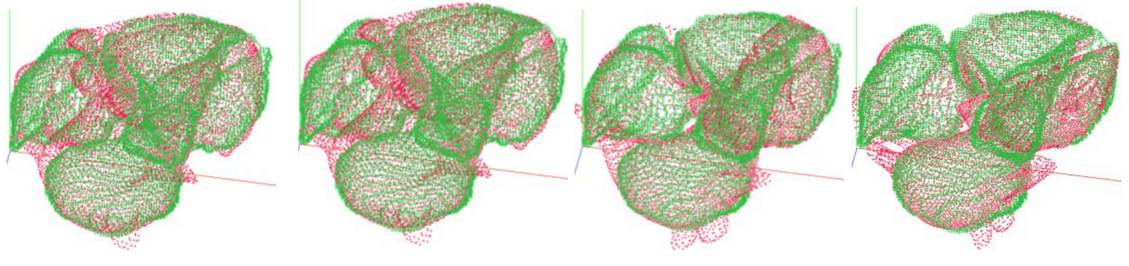}
\centering
\caption{Comparison of different point set registration algorithms. Template's points marked with red dots (moving), data points marked with green crosses (static). All methods were run with $\beta^2=2, \lambda=2, \omega=0.1$. Alignment results starting from left to right: CPD - template structure cannot be observed; ECPD with $\alpha = 10^{10}$- template structure cannot be observed; ECPD with $\alpha = 10^{5}$ -heart structures are partially observable; CCPD - heart structures are fully observable}
\label{fig:Comparison}
\end{figure}

We implemented CPD, ECPD and CCPD in C++ and tested them on Xeon E5-2660 with K40. Efficient GPU version was implemented with CUDA C++. All experiments were conducted using GPU version.
We tested methods listed above with MMWHS-2017 dataset \cite{mmwhs1,mmwhs2,mmwhs3}. This dataset consists of multiple CT and MRI labeled heart scans (figure \ref{fig:mmwhs}). Among other labels, heart ventricles and atriums are landmarked. For the experiment we will be using only these four labels. Experiments were conducted on 20 training CT volumes with corresponding ground-truth segmentation with goal to align STL template with blood cavities surfaces.\newline
After aligning template's ventricles and atriums, whole heart structure can be personalized using resulting transformation. Unfortunately we cannot measure quality of the alignment for neighborhood structures. An example of template personalization shown at figure \ref{fig:personalization}.\newline
The results of template registration are shown at figure \ref{fig:Comparison}. CPD doesn't incorporate any prior information and ECPD uses modified $\Tilde{P}$ matrix where $$\Tilde{p}_{mn} = \begin{cases} 1, & \mbox{if } y_m, x_n \mbox{ share the same cluster} \\ 0, & \mbox{otherwise } \end{cases}$$. Proposed methods shows the best result among others.\newline
For quantitative evaluation two metrics were used: Hausdorff distance and it's cluster variant. Let us write Hausdorff distance as follows:
$$H(A,B) = \max(h(A,B), h(B,A))$$
where $h(A,B) = \max_{a\in A}{\min_{b\in B}{\norm{a-b}}}$, and it's cluster variant as a mean of $H(A(c),B(c))$:
$$HC(A,B) = \frac{1}{C}\sum_{c=1}^{C}{\max{(h(A(c),B(c)), h(B(c),C(c)))}}$$
where $A(c) \subseteq A$ so that $a \in A(c)$ corresponds to cluster $c$.\newline
\begin{figure}[h]
\includegraphics[scale=0.42]{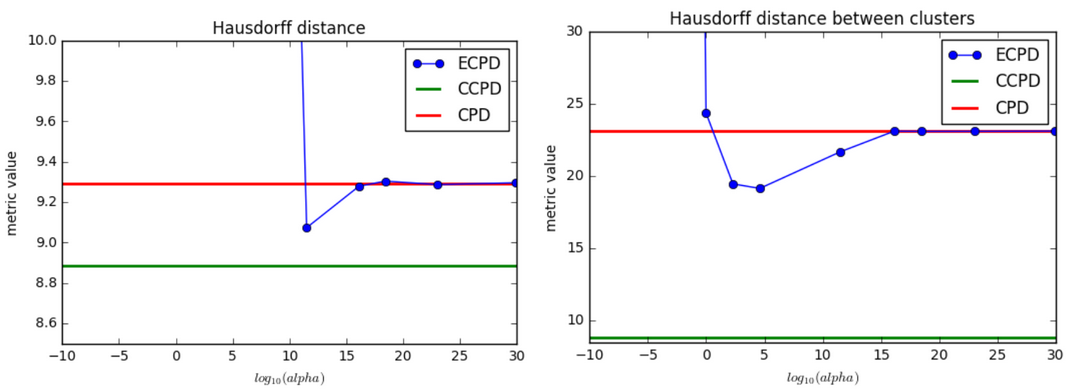}
\centering
\caption{Experimental results. CPD and CCPD doesn't depend on $\alpha$. ECPD doesn't converge for relatively small alphas.}
\label{fig:metrics}
\end{figure}

Table \ref{table:2} shows experimental results. For ECPD column the best result is present. First metric shows quality of alignment with no known information during evaluation stage. The second metric computes average between ground truth clusters, that is more relevant in our case. For both cases proposed method shows better results.

\begin{table}[h]
\centering
\begin{tabular}{||c c c c||} 
 \hline
 Metric & ccpd & cpd & ecpd \\ [0.5ex] 
 \hline\hline
 Hausdorff & 8.88 & 9.29 & 9.07 \\ 
 Cluster Hausdorff & 8.80 & 23.09 & 19.13 \\
 \hline
\end{tabular}
\caption{Comparison of proposed method with ECPD and CPD.}
\label{table:2}
\end{table}

\section{Conclusion \& Discussion}
The problem of non-rigid point set registration is a key problem for many computer vision tasks. In the paper we discussed point set registration problem for clustered point clouds. As an example, points extracted from CT scan of the person's heart can have such property. Heart ventricles and atriums are considered as clusters in this case.\newline
We presented modified version of Coherent Pint Drift for special type of point sets - sets that can be clustered. We defined loss function and showed the way to optimize it using EM algorithm. In the experiments section we compared three methods: proposed one, Coherent Point Drift and Extended Coherent Point Drift that is able to embed prior information on sparse correspondences. We evaluated them in the same environment using MMWHS-2017 data. Proposed method shows more then two times better accuracy than its competitor.\newline
Presented approach is robust to outliers and noise presence and able to produce precise enough transformation even for heart computational models personalization. 

%
%
%
%
%
%
%
%
%
%

\begingroup
\let\clearpage\relax
\bibliographystyle{splncs04}
\bibliography{bibtex}
\endgroup
\end{document}